# Undergraduate Research of Decentralized Localization of Roombas Through Usage of Wall-Finding Software

Madeline Corvin, Johnathan McDowell, Timothy Anglea, and Yongqiang Wang,
*Department of Electrical and Computer Engineering, Clemson University*
Clemson, SC, USA
yongqiw@ clemson.edu

*Abstract*—This paper introduces the research effort of an undergraduate research team in realizing robot localization. More specifically, the undergraduate research team developed and tested wall-following software that allowed a ground robot Roombas to independently find their positions within a defined space. The software also allows a robot to send its localized position to other Roombas, so that each Roomba knows its relative location to realize robot cooperation.

## I. INTRODUCTION

We set out with the goal to create software that would allow for autonomous controlling of Roombas through localization. For our purpose, localization meant the Roomba had the ability to navigate within a closed space. The goal was for it to be able to calculate its own position within the "world" it was located in and then send that information to the machines that are also within the same space. This allows for the Roombas to move freely within their space while knowing where everyone's positions. We wanted the Roombas to be able to communicate to one another so that eventually more complex moving patterns could be created where they are able to follow each other without outside influence.

Throughout this project's lifespan, many different approaches have been attempted to determine the Roomba's location, both relative to the objects around it and to other Roombas. One method involved using sound, or "sound localization," which has been looked into extensively by other researchers and has been implemented in many different ways [1]. In the case of this project, three microphones were placed on the Roombas. The idea was that a sound would play and because the microphones were placed in different places on the Roomba, triangulation could be completed using the time it took for each microphone to activate [2].

This system was never fully implemented and demonstrated errors when tested. Sometimes the Roomba would follow in the general direction, but other times would go off in a random direction. This could have been because the microphones were not picking up the correct sound and thus were following something random. If these problems could have been fixed, a centralized method could be developed where one Roomba plays a sound for the others to follow.

There has also been an attempt to use a compass chip to determine the machine's relative position [3]. When tested within our building, the compass would give back unreliable data because of the relatively strong magnetic fields present within in the building compared to that of the earth's natural magnetic field. As a result, the compass wasn't able to keep track of magnetic north. Other sensors such as gyros, sonar, IR, and wheel encoders have been used to attempt to solve this problem [4]. Issues with data collection and getting consistent and useful data was an issue with these methods as well because the sensors were often very phenoxy.

Unlike other methods tried, our approach to localization for this project did not use sound localization, centralized coordination of the Roombas, or sophisticated sensors like sonar or LIDAR. Instead, we attempted to implement decentralized localization using the sensors (like the bumpers) already located on the Roomba.

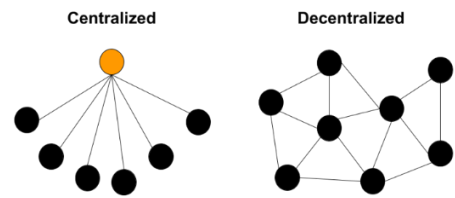

Figure 1: Visual representation of centralized and de-centralized systems

Our decentralized approach meant that no individual Roomba would be in charge. Instead, each Roomba would be responsible for updating the others on its location while simultaneously keeping track of all the other Roombas' locations. A decentralized method allowed for testing on one individual Roomba, since only one was needed for localization and location calculations. By relying on fewer sensors, we were hoping to cut out many of the problems those before us have run into. We also hoped to avoid the more complex math that using other methods (such as triangulation of sound in the case of sound localization) required.

## II. METHODS AND RESULTS

### A. Project Goals

Our goal was to work on having a Roomba find and identify a corner, so that the robot could then use the coordinates of that corner to figure out its location in the world. All code was to be written in Python using the existing RoombaCI_lib.py library in the project's GitHub repository [5].

We made a few basic assumptions to simplify our process. First, we assumed that there would only be one Roomba and no obstacles; this allowed to us to assume that anything the Roomba hit would be a wall (or a corner). We also assumed that the world the Roomba was in was a rectangular room with predefined dimensions known to the Roomba. Finally, we assumed that in our scenario, the Roomba knows its starting orientation and has defined an arbitrary starting position.

B. *Initial Research and Planning*

Before we started writing code, we did some research into the Roomba's bumpers, light bumpers, and wheel encoders, as we expected to use these sensors extensively in corner-finding.

The Roomba's front bumpers were perhaps most important, as the way the bumpers detected and reported hits impacted how we structured our logic in the code later on. As there are two bumpers on the Roomba, there are four possible situations for the Roomba to detect: a left bump, a right bump, a front hit (where both bumpers are hit), or no bump (where neither bumper is hit). The Roomba's documentation [6] indicates that the data from the bumper is stored as an unsigned byte, structured as shown below:

Figure 2: Bumper byte documentation

So, a hit to the left bumper would be stored as 00000010 (decimal 2), a hit to both bumpers would be stored as the byte 00000011 (decimal 3), and so on. As stated above, this information was used to structure our conditionals and logic.

The documentation also indicated that the information from the bumpers could be requested from the Roomba by asking for sensor Packet 7. (We used the StartQueryStream() and ReadQueryStream() functions in the library to do this.) The fastest this data could be requested was approximately once every fifteen milliseconds.

The Roomba also has encoders on each wheel. Our plan was to use the wheel encoder counts to help with corner identification, so we also looked into how that data was stored. The documentation indicated that the wheel encoder counts were stored as 16-bit signed bytes, structured as shown below:

Figure 3: Left wheel encoder documentation

Figure 4: Right wheel encoder documentation

This data could be requested from the Roomba by requesting Packets 43 and 44 in the same way as the bumper sensor data. Note that the data in these packets would be raw values, meaning they would need to be converted to different units (such as millimeters) to actually be useful. Various test programs in the GitHub already make use of the encoder data, so we planned to use those for reference when writing our code.

We split our project into several stages to make working on it easier. These stages are as follows:

1. Initiating movement and identifying when a wall has been hit
2. Turning once a wall has been hit (either clockwise or counterclockwise depending on which bumper was hit)
3. Wall-following
4. Identifying when a corner has been reached
5. Identifying which corner has been reached and updating position using the corner's coordinates

C. *Implementation*

To begin with, we wrote a program called Roomba_WallFinder_Test.py. (All of our code for this project is contained within this program.) This program's function was to wake and initialize the Roomba, continuously request the bump sensor packet (Packet 7) from the Roomba in a query stream, and report the results. The value of the bumper byte and the corresponding type of bump (left, right, both, or none) were printed to the screen for testing purposes.

```
00000000, 10183, -4084
00000000, 10186, -4081
00000000, 10191, -4077
00000011, 10194, -4073
00000011, 10198, -4069
00000011, 10199, -4067
00000011, 10197, -4069
00000011, 10196, -4071
00000001, 10193, -4073
00000001, 10190, -4075
00000000, 10187, -4078
00000000, 10184, -4081
00000000, 10181, -4084
00000000, 10178, -4087
00000000, 10175, -4090
```

Figure 5: Sample bumper and encoder data

In the data above, the bumper byte is printed first and then the raw left encoder wheel count followed by the raw right wheel encoder count. (This data was sampled at the highest rate possible.) Note that the Roomba typically reported a bumper byte of 00000000, meaning there was no bump. The occasions where the bumper byte was non-zero (i.e. 00000001) indicate a point in time where one or more of the bumpers were hit. This data makes sense with what we would have expected based on our research. (To determine which bumper was hit, we modulated the bumper byte received from the Roomba by four in our conditionals. This means that a right hit would result in a 1, a left hit in a 2, and so on.)

This program's purpose was to allow us to test our understanding of the bump sensors and bumper logic, not to mention our understanding of the query stream functions, without having to deal with the hassle of a moving Roomba. As stated, the program written in this stage became the framework we built on for the rest of the project. We did eventually add in a move command and tested the bumper logic that way as well.

Our next step was to have the Roomba actually navigate a wall hit. The general idea was to have the Roomba execute a "bump-and-bounce" maneuver: move back slightly, turn slightly, move forward, and repeat until the obstacle is cleared. The bump-and-bounce is controlled using a series of timers, as depicted in the code below, most of which was recycled from Roomba_BumpControl_Test.py in the library.

```
# initialize timers
#spinTime = (235 * math.pi) / (4 * spnspd) # from formula
spinTime = 0.75 # arbitrary number
backTime = 0.25
dataTimer = time.time()
timer = time.time()
moveHelper = (time.time() - (spinTime + backTime))

#timer for the backward movement, then the spin
if (time.time() - moveHelper) < backTime:
        Roomba.Move(moveVal, 0) # backward movement
elif (time.time() - moveHelper) < (backTime + spinTime):
        Roomba.Move(0, spinVal) # spin
else:
        Roomba.Move(movSpd, forwardSpin) # forward and spin
```

Figure 6: Roomba movement timing code

In the code above, the timers were contained in the variables moveHelper, backTime, and spinTime. These timers were set to arbitrary values during initialization and were used in conjunction with the Python time library. Decreasing the amount of each timer decreased the amount of time the Roomba could turn or move forwards, so we were able to use them for optimizing the Roomba's movements. However, we had to be careful not to decrease the timers too much, or the Roomba would barely turn or move at all. Again, no specific formula for determining the timing was followed and the values of the timers remain arbitrary.

The direction that the Roomba turns while trying to navigate the wall is determined by which bumper was hit. In the case of a left bumper hit, the shortest rotation for the Roomba is in the clockwise direction; the opposite would be true for a right bumper hit, where the shortest rotation is in the counter-clockwise direction. This is true regardless of which wall is hit, so for efficiency we opted to have the Roomba turn the same direction each time for a certain type of bumper hit. (There is also the fact that the bump-and-bounce maneuver would probably fail if the Roomba executed the larger turn, as it might end up drifting away from the wall.)

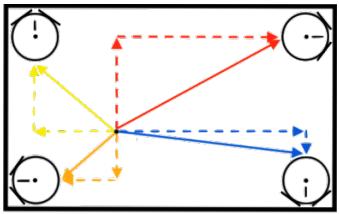

Figure 7: Roomba turning diagram for left bumper hit

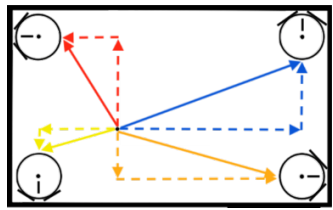

Figure 8: Roomba turning diagram for right bumper hit

Rotation was controlled using the second parameter, y, of the Move command from the library. The input x controls the forward speed of the Roomba, so its sign needs to be positive in order to have the Roomba execute a forward turn. And as noted in the comments for this function, a positive value of y will result in a clockwise rotation, while a negative value of y will result in a counter-clockwise rotation.

```
''' Send command to Roomba to move
        Parameters:
                x = integer; common wheel speed (mm/s); x > 0 -> forward motion
                y = integer; diffential wheel speed (mm/s); y > 0 -> CW motion
        Error may result if |x| + |y| > 500. '''
def Move(self, x, y):
        RW = x - y # Right wheel speed
        LW = x + y # Left wheel speed
        self.conn.write(b'\x91') # Send command to Roomba to set wheel speeds (145)
        self.conn.write((RW).to_bytes(2, byteorder='big', signed=True))
        self.conn.write((LW).to_bytes(2, byteorder='big', signed=True))
```

Figure 9: Implementation of the Move command from RoombaCI_lib.py

To make use of this, when the type of bumper hit was identified, we assigned a value (positive or negative) to a variable spinVal; this variable was then used for the second parameter in the Move command, executed in the timing sequence. This means that for a left bumper hit (which requires a clockwise rotation), spinVal was set to be 100, whereas for a right bumper hit it was set to be -100. (The actual value of spinVal is arbitrary.) For a front hit, we made the spin direction random, as the Roomba would have to turn the same amount in either direction. This means that the value of spinVal also needed to be randomly set to either positive or negative.

```
# both - front hit
print("Both bumpers hit!")
y = random.randint(0,1)
spinVal = random.randint(spnspd - 50, spnspd + 50)
if y == 0:
        spinVal = -spinVal
moveVal = -100
```

Figure 10: Front hit code

We also made use of the Move command for our next stage, wall-following. The idea was to have the Roomba follow along the wall to find a corner, once it has cleared the bump-and-bounce stage. However, it was necessary to account for the fact that the Roomba would probably drift away from the wall and into the room (depending on its orientation when it finished clearing the wall).

We briefly considered using the light bumpers to track how far way the Roomba drifted from the wall, but this method became too complicated for us to work out in our limited time. Instead, we opted to have the Roomba turn back towards the wall in its forward movements, as illustrated in the diagram below.

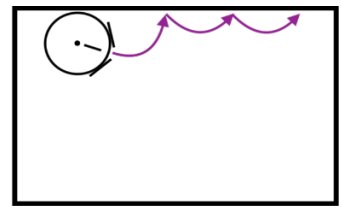

Figure 11: Diagram of Roomba wall-following motions

This movement was created using a variable called forwardSpin, which was assigned to be the negative of whatever spinVal was. (Note that since the value of spinVal causes the Roomba to turn away from the wall, the negative of spinVal will cause the Roomba to turn back towards it.) We somewhat optimized the Roomba's wall-following motions by changing the timers and the spin speed, similar to what we did for the wall-clearing maneuver.

The last steps in our plan were to work on identifying when the Roomba has found a corner, determining the Roomba's position in the world, and then having the Roomba report its position to other Roombas for later use in group localization. We were not able to complete any of these stages this semester. However, we did develop a plan for corner identification involving the Roomba's wheel encoders. When we tested our wall-following code in the lab, we observed that the Roomba finds a corner while wall-following, it gets "stuck" in that corner. This means that the Roomba stays in approximately the same position but tends to alternate back in forth between which bumper is hit (i.e., left, right, left, right, etc.).

We planned to have the Roomba save its position (via the encoders) each time it hits a wall in addition to saving which bumper sensed the collision. If the Roomba noticed an alternating bumper hit pattern (such as LRLR or RLRL) and also noticed that it stayed within a certain distance of its previous position each time, then it would be programmed to assume that it had found a corner. Note that this plan does not really take into account front hits, as we assumed that the Roomba was unlikely to hit the corner head-on.

Once the Roomba had determined it was in the corner, we intended to have it use the encoders to work out its new position, perhaps by comparing the horizontal and vertical distance it had traveled to its initial position in the world. For example, if the Roomba's wheel encoder counts indicated it had moved a net positive amount in the vertical direction and a new positive in the horizontal direction, it could be assumed that the Roomba was in the top right corner. (This plan would require the Roomba to define the world in terms of a coordinate system, however, and also may not be easy to generalize to other types of worlds later.) The Roomba would then report its new position using the Xbee to any other Roombas in the world.

As mentioned above, we ran out of time to work these ideas into our code.

D. *Testing*

We set up a test scenario to confirm the Roomba's ability to accurately in find the corner of a room based on what we had already programmed. This test scenario is illustrated in the figure below.

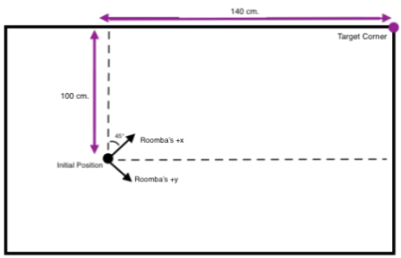

Figure 12: Roomba test scenario

As shown in the figure above, the Roomba was placed in a random starting position in the room at an approximately 45° angle with what we defined to be the true x-axis. We measured the horizontal and vertical distances from this starting position to what we defined as our "target corner" in the room and found these distances to be 100 cm. and 140 cm. (or 1000 mm. and 1400 mm.) respectively. As a result, we set the target corner's coordinates to be (1000 mm., 1400 mm.), as the Roomba would define its starting position to be (0, 0).

Roomba_WallFinder_Test.py was then run and data from the Roomba's wheel encoders was collected every 15 ms. This wheel encoder data was then converted into the Roomba's x- and y-positions in millimeters and plotted using Excel to display the Roomba's path in the room.

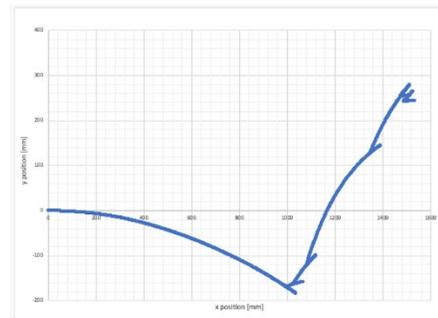

Figure 13: Roomba's path

Based on this data, the Roomba reported its final position in the room to be (245.85 mm., 1494.89 mm.).

However, note that the Roomba defines the positive x-direction to be its initial direction of forward motion, which is not how we defined our coordinate system when setting the coordinates of the target corner. To correct this, we applied the rotation matrix below that rotated the Roomba's position data 45° in the counterclockwise direction.

$$Rv = \begin{bmatrix} \cos\theta & -\sin\theta \\ \sin\theta & \cos\theta \end{bmatrix} \cdot \begin{bmatrix} x \\ y \end{bmatrix} = \begin{bmatrix} x\cos\theta - y\sin\theta \\ x\sin\theta + y\cos\theta \end{bmatrix}$$

Figure 14: Rotation matrix

This aligns the Roomba's x-position data with what we defined to be the x-axis (and additionally, the Roomba's y-position data with our y-axis). The rotated position data was also plotted in Excel.

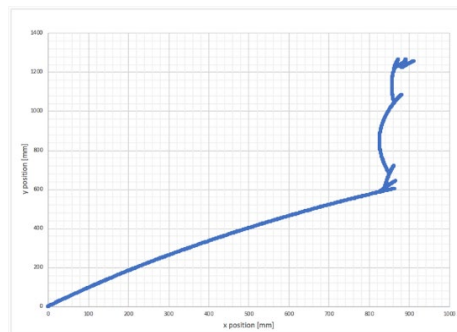

Figure 15: Roomba's path with rotation matrix applied

Using the rotation matrix, the Roomba's final reported position is (883.39 mm., 1230.70 mm.). Note that this position is measured with respect to the center of the Roomba, meaning that the position is offset by 117.5 mm., or the length of the Roomba's radius (taken from the documentation). When the radius is taken into account, the Roomba's final reported position becomes (1000.89 mm., 1348.20 mm.), which is very close to the target corner's position.

This same test was run a few more times and the result each time was found to be approximately the same. As mentioned earlier, at this point the Roomba would need to work out where its actual position in the room, since its starting position was not actually (0 mm., 0 mm.).

## III. CONCLUSIONS AND FURTHER WORK

All of the work we completed has been tested and works without error. However, there is some work that still needs to be done and some things that could be tried to further optimize this project.

### A. *Corner-Identification and Reporting Position*

As mentioned, we did not complete the corner-identification or coordinate-reporting stages of our project. In order to be considered complete, these stages would need to be written and tested. Corner-identification could potentially be done using the encoders and the plan laid out in the previous section. (This plan may need to be extended to consider front-hits.) The Roomba could also use the wheel encoders to determine its new coordinates once it finds the corner, as mentioned.

We also considered a plan that involves using an Inertial Measurement Unit to determine the Roomba's new coordinates; if this idea were to be used, a section of code would need to be added to the beginning of Roomba_WallFinder_Test.py that would initialize the IMU. (The place where this would go is indicated with a comment in the program.) The math for using the IMU data (i.e., acceleration) to calculate the new coordinates would also need to be worked out. We did not create a specific plan for either possibility.

The Roomba also needs to be able to report its coordinates to other Roombas once it finds a corner. Implementing this involves determining what data type(s) the Roomba's message contains (i.e., would the coordinates be sent as strings? Would they be sent as integers?) as well as working out the timing of sending and receiving the messages. Once this reporting stage is completed, the Roomba(s) can be programmed to execute specific coordinated maneuvers such as "follow the leader" or forming a platoon. This was our end goal.

### B. *Generalization*

We made a few assumptions at the beginning of this project to simplify the implementation. As stated, these assumptions include: the world is rectangular and has dimensions known to the Roomba; there is only one Roomba in the world; and the Roomba knows its starting direction and has defined its starting position. Making these assumptions create a specific case that is neither likely (as most rooms have obstacles and are not simple rectangles) or useful (as having only one Roomba in the room defeats the purpose of studying swarm robotics). As a result, our program will need to be generalized so that it can operate in cases that do not fit into our assumptions.

For example, there will probably be more than one Roomba in the world in most cases. This means that if the Roomba's bumpers detect a collision, that collision could be with another Roomba instead of a wall or corner. The Roomba must be able to distinguish between hitting another Roomba and hitting a wall.

One possible solution we considered was having the Roomba stop and check whether the obstacle it hit moves after a certain amount of time. If it moved, it was another Roomba, and if not, then it was likely a wall. This idea breaks down if there are stationary obstacles in the room in addition to other Roombas. As there will probably be obstacles in the world (like furniture or stray objects), the Roomba will also need to distinguish between hitting a wall, hitting another Roomba, and hitting a stationary obstacle.

It could also be useful to have the Roomba remember where certain obstacles were located in the world so that it can avoid hitting them again later. The Roombas also could share this kind of information between themselves to create a more detailed map of the world.

It is also important to consider that in general, the Roombas will probably not know anything about the world like they do in the scenario we used. Even if the world is still rectangular in shape, as it was in our project, the Roomba likely will not know the dimensions beforehand. And it is likely that the room will not be rectangular in shape; for example, it could have five corners or be circular and have no corners at all. These things could impact the Roomba's ability to update its coordinates once it finds a corner, depending on how that stage gets implemented.

A solution that allows the Roomba to determine its coordinates in every situation to be found to make our approach to localization more general.

### C. *Other Ideas*

There are few miscellaneous ideas relating to this project that may be useful to try. Firstly, while we used timers to coordinate the Roomba's different movements in our program, we also discussed the possibility of using states, such as those illustrated below.

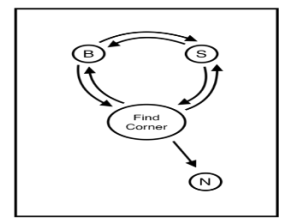

Figure 16: Roomba state diagram

Possible states for the Roomba to be in include "backwards," "spin," or "find corner" (which is where the

Roomba would be if it is not attempting to clear a wall or in a corner). Note that this list of states is not all of the possible states we could define, and we may wish to create more. A variable could be used to track which state the Roomba is in; for example, making the variable equal to one for "spin". Using states rather than timers could be beneficial, as it might be more efficient both in code (requiring fewer lines) and in terms of the Roomba's actual movements. States may also be easier to generalize than our current solution with the timers.

It also might be useful to look into using the Roomba's light bumpers, rather than its actual front bumpers, to implement wall-following. Our current solution using the bumpers works, but it is a bit rough and inefficient. If a solution could be found that would allow the Roomba to follow the wall without having to hit the wall repeatedly (and as result, waste time executing the wall-clearing maneuver), a significant amount of time could be saved.

The light bumpers could be used for this purpose by alerting the Roomba when it has moved a certain distance away from the wall. The Roomba would probably determine this based on the intensities received from the light bumpers. However, it is important to keep in mind that while we might increase our time efficiency using this method, we might lose efficiency in terms of code.

Figure 17: Light bumper documentation

The reason for this is because several factors make the implementation more complex, as we found out during our attempts to use the light bumpers. As shown in the documentation above, the Roomba has several more light bumpers than physical bumpers (five compared to two). This means that there are more possible situations to consider when using the light bumpers than there are with the physical bumpers, where there are only four (no hit, left hit, right hit, front hit).

It is also possible to receive the light bumper data in different formats. If Packet 45 (shown above) is requested, then a condensed, binary version of the light bumper data will be received, with each light bumper being represented by a bit. More detailed information from each light bumper (i.e., the specific intensities) can also be requested. For example, the strength of only the front right light bumper signal can be obtained by requesting Packet 47. This data will be received as an unsigned, 16-bit value as stated in the documentation below.

Figure 18: Light bumper (front left) documentation

This could be useful if the Roomba were attempting to use the light bumpers to follow along a wall on its left side. Similar data for the other light bumpers can be obtained by requesting Packets 46 (left), 48 (center left), 49 (center right), 50 (front right), and 51 (right). As with the wheel encoders and the bumpers, this would be done using the QueryStream functions from the GitHub library.

This solution could be better than our current one but would require more research and testing of the Roomba's light bumpers than we were able to complete.

If the current wall-following solution is kept, it probably should be optimized further. The current spin times and forward speeds were selected to be arbitrary values and could be adjusted to make the Roomba's movements more efficient. For example, limiting the amount of time that the Roomba has to turn away from the wall will decrease the amount of time that it has to take turning back, which will make the Roomba's overall forward progress faster. Other variables, like the Roomba's forward speed or turning speed could be increased or decreased to help with this as well.

Finally, once the Roombas are capable of localization in all (or at least most) cases, code will need to be written that tells them to execute coordinated maneuvers, as this was the end goal of developing localization in the first place. For example, once all of the Roombas have reported their coordinates, they could be commanded to all report to the center of the room. Note that they should be able to determine how to reach the center based on the dimensions of the room and their current coordinates. (The math for this will need to be worked out and then written into code.)

Then, once all the Roombas are in the center, they might be programmed to perform a synchronized "dance," or play follow-the-leader, or form platoons. The necessary implementation of anything like this will vary depending on what type of maneuver is desired.